\documentclass{article}

\usepackage[preprint]{neurips_2023}

\usepackage[utf8]{inputenc} 
\usepackage[T1]{fontenc}    
\usepackage{hyperref}       
\usepackage{url}            
\usepackage{booktabs}       
\usepackage{amsfonts}       
\usepackage{nicefrac}       
\usepackage{microtype}      
\usepackage{xcolor}         
\usepackage{graphicx}

\usepackage{natbib}
\title{Coherent Wave Dynamics and Language Generation of a Generative Pre-trained Transformer}

\author{Tao Hong \\
  Meta Platforms Inc. 
  \\ One Hacker Way, Menlo Park, CA 94025\\
  \texttt{taohong01@gmail.com} \\
}

\begin{document}

\maketitle

\begin{abstract}

Large Language Models (LLMs), such as the Generative Pretrained Transformer (GPT), have achieved tremendous success in various language tasks, but their emergent abilities have also raised many questions, concerns, and challenges that need to be addressed. To gain a better understanding of the models' inner mechanisms, we analyze the hidden state and channel wave dynamics in a small GPT, focusing on the coherence of wave patterns in terms of cross-channel correlation and individual auto-correlation. Our findings suggest that wave dynamics offer stable, consistent, and repeatable intrinsic oscillation modes, along with context-aware plasticity and expressiveness in language generation. By analyzing wave patterns, coherence, and clustering, we provide a systematic way to identify and interpret the functionality of the hidden state channels, paving the way to understand and control higher-level language pattern formation. In addition, we investigate the Poisson statistics of spelling errors in text sequence generation across various levels of model training and observe a phase-transition-like process. As coherence builds up, there is a competition between the generation of correct and misspelled words. However, once the model is adequately trained and significant coherence has emerged, the coherent process becomes strong enough to effectively suppress spelling errors, preventing the cascade amplification of defects. Interestingly, the distribution of correct spellings transitions from Poissonian to Sub-Poissonian, while the distribution of misspellings shows the opposite trend. By leveraging concepts and techniques from quantum physics, we gain novel insights into the dynamics of the small GPT. This approach can be extended to larger language models that exhibit more complex coherent language patterns, opening up opportunities to interpret their emergent capabilities and develop more specialized models.

\end{abstract}

\section{Introduction}

Large language models (LLMs) have demonstrated impressive abilities, surpassing people's expectations with their emergent capabilities~\cite{instructgpt,gpt,gpt2,gpt_2,gpt3,openai2023gpt4,chowdhery2022palm,jason2022}. However, the lack of in-depth understanding of their inner workings and technical control over these large models has raised concerns and questions. The sheer size of these models makes training and maintaining them extremely challenging, and finding better solutions to make them smaller and more manageable is critical. However, it is not yet clear how to create smaller models tailored to specific tasks without sacrificing significant functionality. Despite people's frequent interaction with LLMs, their empirical understanding mostly hovers on the surface. The ambiguity and lack of theoretical support surrounding the emergent ability continue to spark controversy. In reality, while people are excited about the models' long sequences of logical reasoning capabilities, they are often surprised by the models' inability to distinguish real information from fake or to provide sensible responses. There is currently no systematic approach to ensure that these large models master common sense. Deepening our understanding of LLMs' operating mechanisms is crucial to address the concerns, questions, and challenges that have arisen due to their emergent abilities. Only by gaining a solid understanding of these models' inner workings can we establish human control over them and develop confidence in their ability to operate in a safe and ethical manner.

Our aim in this study is to deepen our understanding of LLMs by analyzing the hidden state wave dynamics in a small GPT. Specifically, we will focus on the coherence of wave dynamics in terms of cross-channel correlation and individual auto-correlation and investigate its impact on Poisson statistics of language patterns. We will leverage wave patterns, coherence, and clustering to demonstrate a systematic way of identifying and interpreting the functionality of the hidden state channels, paving the way to understand and control higher-level language pattern formation. Additionally, we will investigate the Poisson statistics of spelling errors in text sequence generation across various levels of model training. Our findings will reveal a phase-transition-like process that propagates the wave dynamic coherence to the language pattern coherence, which could potentially explain the emergent capabilities when a model scales up. Through our study, we hope to shed light on the inner workings of LLMs, address the concerns and questions that have arisen due to their emergent abilities, and pave the way for a more efficient and effective use of these models in a safe and ethical manner.

\section{Related Work}

LLMs have shown impressive abilities in performing complex reasoning through prompt-guided learning, leading to improved performance on various tasks~\cite{jason2022}. However, the underlying mechanism behind these emergent abilities is still unclear. A recent explanation proposes that the researcher's choice of metrics may infer emergent abilities rather than fundamental changes in the model's behavior~\cite{schaeffer2023emergent}, but it does not rule out the possibility of higher-order phase transition, which may not exhibit evident discontinuity.

Despite sharing a similar neural network architecture\cite{ashish2017}, the underlying mechanisms and information distribution of LLMs remain complex and unclear. For example, a study on autoregressive transformer language models found that direct manipulation of computational mechanisms through Rank-One Model Editing (ROME) is a feasible approach for model editing, enabling the exploration of new possibilities and architectures~\cite{meng2022massediting,meng2023locating}.

Coherence in language and coherence in physics are two distinct concepts, but there are connections between them. Coherence in language refers to the consistency and logic of a text, while coherence in physics refers to the relationship between waves. Some neuroscience research suggests that coherence in language can be correlated with coherence in brain electrical signals~\cite{weiss2003}. However, direct evidence and understanding of how physics coherence translates into language coherence are still lacking.

In this paper, we propose a novel approach by applying concepts and methods from quantum physics to analyze and interpret the language model through the lens of wave dynamics. Our aim is to bridge some of the gaps in our current understanding of language coherence formation and emergent abilities and provide new insights into these areas.

\section{Model Architecture}

To simplify our discussion and focus on the fundamental dynamic properties of GPT models~\cite{gpt,gpt2,gpt_2,gpt3,openai2023gpt4,instructgpt,rlhf}, we will use a smaller version of the model that retains its essential architecture while excluding unnecessary complexity. While we acknowledge that some emerging capabilities of LLMs may strongly depend on model size and training scale, we believe that the smaller model we use will still be sufficient to reflect the basic properties and provide insights into trends and possibilities.

The model we will use is called minGPT, which is a character-level transformer-based language model trained on a small dataset called tiny Shakespeare~\cite{andrej2022}, composed of a sequence of 1,115,394 characters. It models how the characters in the dataset follow each other, producing text that resembles Shakespeare's writing style.~\cite{andrej2022}

MinGPT is essentially the decoder part of the original transformer~\cite{ashish2017}, but with a small variant in pre-layer normalization~\cite{ruibin2020}. It has an embedding dimension of $192$ for $65$ unique character tokens and consists of $8$ hidden layers with $6$ heads of attention and ~$3.6$ million parameters. The context length varies from $1$ to the maximum given by the block size, which is $128$. During training, we used a dropout rate of $0.2$, a learning rate of $1.0\times10^{-3}$, and a batch size of $64$ independent character sequences. The model was trained on the tiny Shakespeare dataset at the character level, which contained $1,003,854$ character tokens in the training set and $111,540$ character tokens in the validation set. The model training has been stopped at 3 different conditions, as shown in Table~\ref{tab:model_training_levels}.
\begin{table}
    \caption{Model Training Stop Conditions}
    \label{tab:model_training_levels}
    \centering
    \begin{tabular}{llll}
    \toprule
         Condition  & Epoches     & Training Loss & Validation Loss  \\
         \midrule
         1          & $6500$    &  $1.11$       & $1.49$  \\
         2          & $1000$    &  $1.45$       & $1.63$ \\
         3          & $100$     &  $2.45$       & $2.47$  \\
    \bottomrule
    \end{tabular}

\end{table}

For our analysis, we generated $20,000$ character tokens using a $1\times1$ zero tensor to kick off the generation. Before generating each token, we followed the time sequence to retrieve the intermediate hidden states, which act as an input to the first layer norm of each attention block, through all the $8$ blocks. In the next section, we will present our analysis of these hidden states.

\section{Hidden State Pulsation and Word Generation}

In general, complex oscillating pulsations are observed in the hidden states. To unravel these complexities, we can examine a simpler case by sending a single space character to trigger the network to start token generation. As shown in Fig. 1(a), the initial hidden state at the input of hidden layer HL0, produced by the token embedding and the position embedding, starts with very little ripple and undergoes rapid amplification through eight hidden layers. This amplification occurs through the self-attention mechanism because there are no other tokens in the context. The amplification selectively stretches out the amplitude span on both directions for some channels, resulting in a significantly changed waveform across 192 channels at the last layer HL7 compared to that at the first layer HL0. To quantify the change, we calculate the cosine similarities of these waveforms from HL0 through HL7 with respect to HL0, resulting in similarity scores of 1.0, 0.87, 0.76, 0.72, 0.66, 0.62, 0.61, and 0.57.

To investigate how this kind of amplification occurs in the presence of more context tokens, we captured a snapshot of the amplitude evolution of a sequence of context tokens while they were producing a prediction for the next token, as shown in Fig. 1(b). We observed multiple space tokens in this sequence, all of which exhibited significant spikes compared to other tokens. Channel 14 and 129 seemed to be signature channels for space tokens. However, we also saw differences in these spikes due to differences in their surrounding context tokens as well as their positions.

Furthermore, we observed that the amplitudes of channels 14 and 129 often swing in opposite directions, suggesting that their interplay can introduce an interference effect through downstream layer connections, akin to complex wave functions in quantum mechanics in terms of the two degrees of freedom: either the real part and the imaginary part or the amplitude and the phase. This interference can produce either constructive or destructive interference and thus control the probability of token generation in the final output. This dynamical process enhances each token's plasticity and expressiveness for generating complex correlated language patterns.

We also investigated the consistency of the waveforms generated by each token under different contexts. To illustrate this, we collected waveforms generated by the same word and then computed their means and standard deviations. As shown in Fig. 2, we observed stable wave patterns associated with each character token in two similar words, $'you\ '$ and $'your'$. Specifically, we averaged the oscillation amplitudes over 54 instances of $'you\ '$ and 37 instances of $'your'$ that we found in the continuously generated 20,000 tokens. Fig. 2(c) shows a standard deviation much smaller than the corresponding amplitude mean in Fig. 2(b). The consistent wave patterns in the intermediate hidden state waveforms suggest that they are eigenmodes generated by the token embedding vectors, although there are some variations on top of them when the context changes. They could be used for complex waveform decomposition and analysis, for example, helping us to understand how token interactions lead to stable long-range coherent sequences, such as words, phrases, and sentences, etc.

\begin{figure}[htbp]
  \centering
  \includegraphics[width=0.7\linewidth]{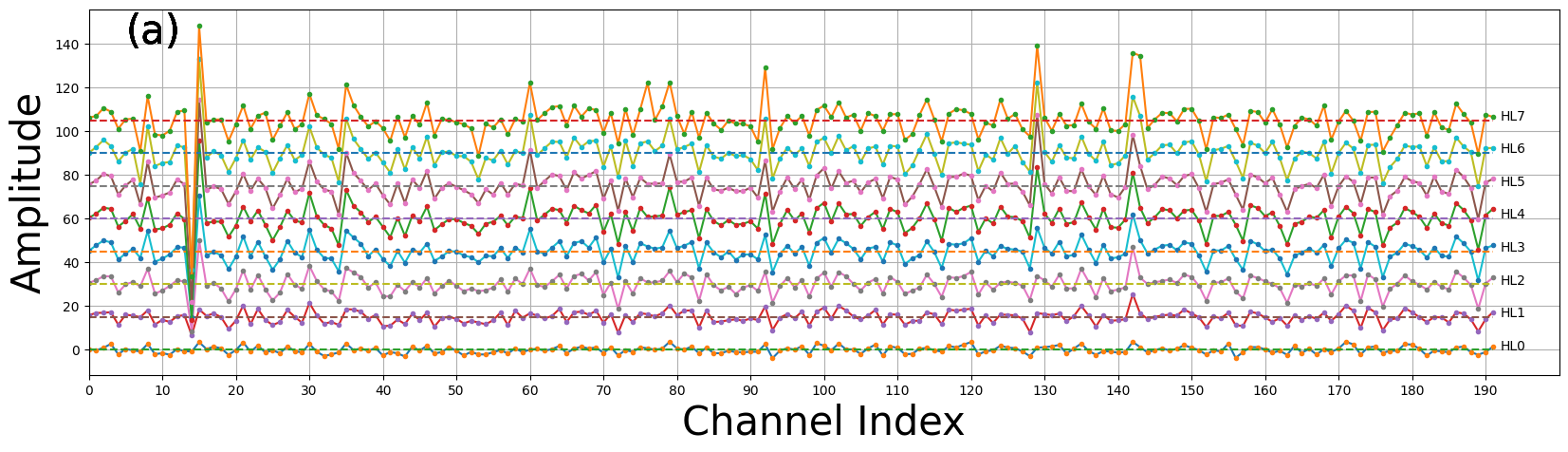}
  \includegraphics[width=0.7\linewidth]{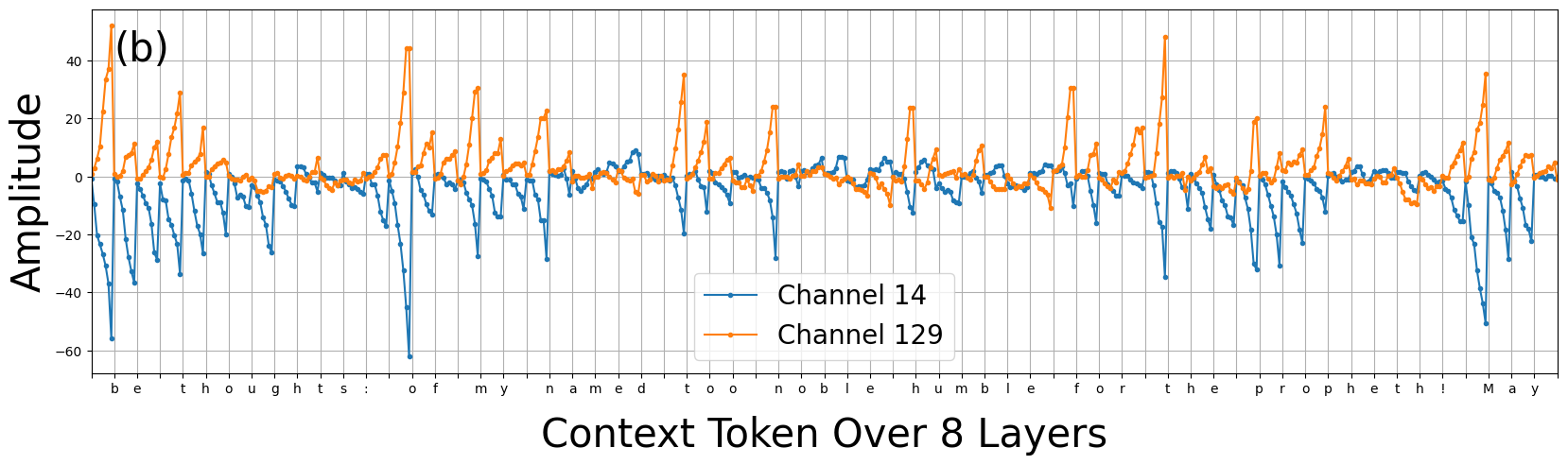}
  \caption{ Hidden state waveforms and amplitude evolution through the network. (a) shows the waveforms across 192 channels and their amplitude evolution through 8 hidden layers when a single space token triggers the network. Each dot represents the amplitude of a channel, and solid lines connect them for visualization. The labels on the right, such as HL0, denote the corresponding hidden layer index. Dashed lines through the amplitudes indicate the corresponding zeros, which are evenly shifted upward for ease of observation. (b) illustrates the amplitude evolution at channel 14 and 129 when there are 128 tokens in the context of self-attention. The horizontal axis is labeled with a portion of tokens in the context, and each token is listed under each tick. The corresponding amplitude evolution through eight hidden layers is depicted by eight dots in the following cell.}
  
  \label{fig1}
\end{figure}

\begin{figure}[htbp]
  \centering
  \includegraphics[width=0.7\linewidth]{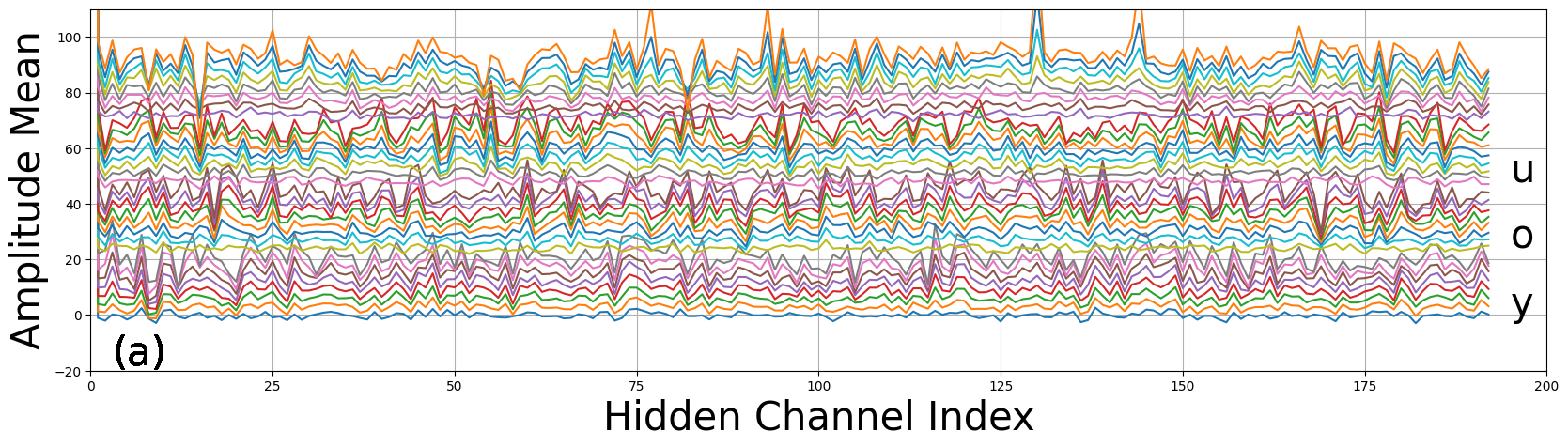}
  \includegraphics[width=0.7\linewidth]{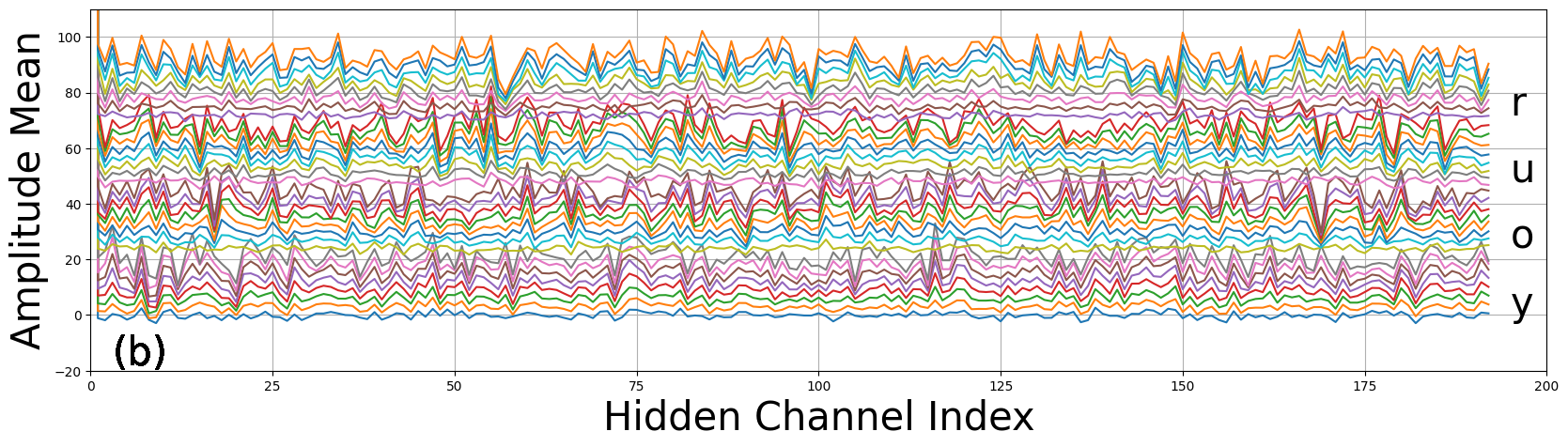}
  \includegraphics[width=0.7\linewidth]{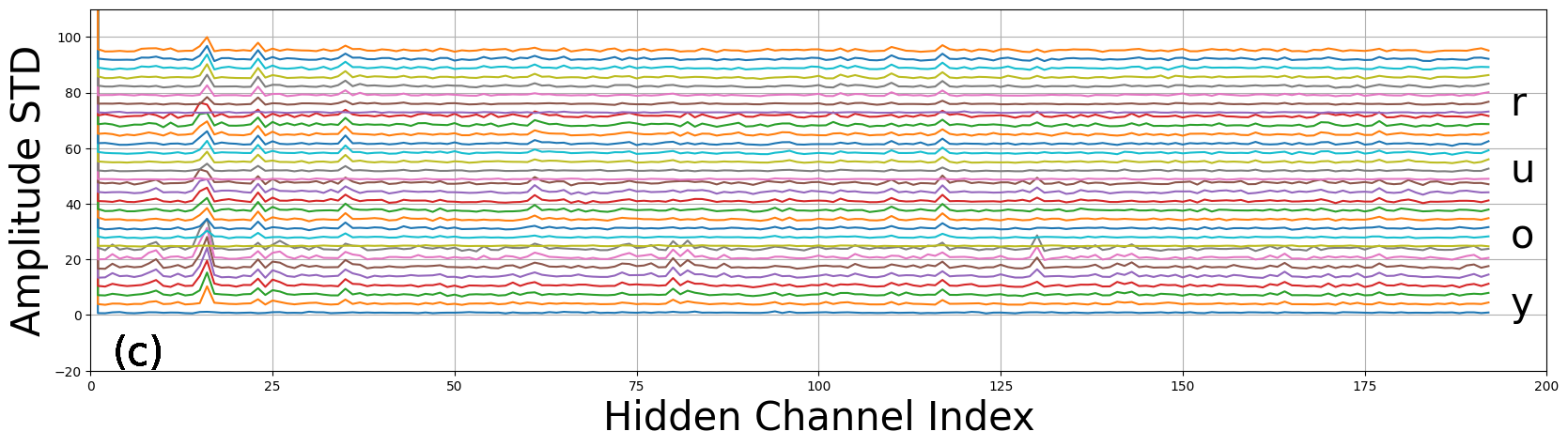}
  \caption{Amplitude Mean and Standard Deviation (STD) for the words $'you\ '$ (a) and $'your'$ (b, c). The solid lines connect the oscillation amplitude data points for each channel, which have been evenly up-shifted to aid in observation. Dots have been omitted to reduce complexity. On the right side of the figure, each character label corresponds to the 8 hidden state waveforms in sequence over time. Note that in (a), there is a space character above $'u'$.}
  \label{fig2}
\end{figure}

\section{Coherence across Hidden Channels}

The stable waveform patterns and synchronization between different channels, as observed in channels 14 and 129, suggest that correlations play a significant role in the hidden state dynamics for specific tokens. To better understand how these channels connect and work collectively in a general case, we calculated the oscillation amplitude correlation matrix between hidden channels. This matrix is analogous to the density matrix in quantum physics and can reflect the system's inherent coherence underlying the dynamics.

Specifically, the correlation is given by $C = \Sigma_{i=1}^{128\times156\times8} A_{ij} \times A_{ik}$, where $A_{ij}$ and $A_{ik}$ represent the oscillation amplitudes at channels $j$ and $k$, respectively. The index $i=(s\times128+t)\times8+l+1$, where the snapshot index $s\in[0,155]$, the context token index $t\in[0,127]$, and the hidden layer index $l\in[0,7]$. Since all amplitudes are real numbers, the square density is symmetric about the diagonal, so we only need a sub-triangle of the square matrix.

Fig. 3 shows that significant coherence is built up after training, with hidden channels associated with spaces and other characters controlling common language patterns standing out prominently from the rest. These channels appear as spikes, with very bright dots for positive values or very dark dots for negative values in the heat map of Fig. 3(a). They are relatively sparse and distributed on the wings, as shown by the histogram on the right in Fig. 3(b). We also calculated this type of correlation matrix for single character tokens and can identify signature channels with high correlation matrix elements for each token or string pattern. However, given the complexity, we will defer this analysis for future publication.

\begin{figure}[htbp]
  \centering
  \includegraphics[width=0.7\linewidth]{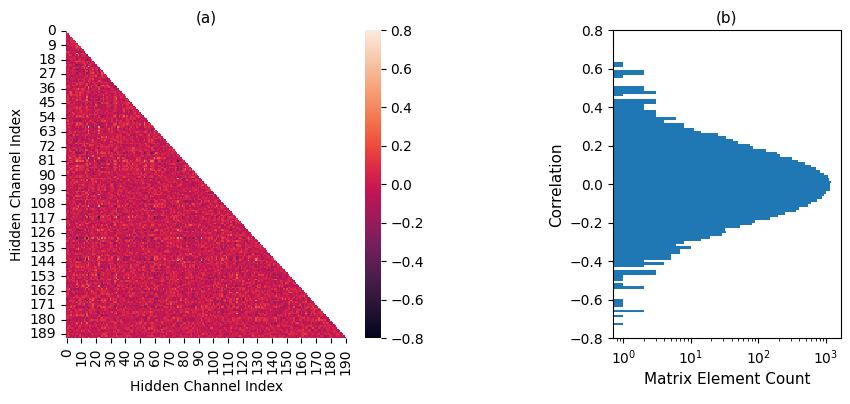}
  \caption{Coherence across hidden state channels. (a) shows the triangle correlation matrix and (b) shows the histogram of correlation matrix elements.}
  \label{fig3}
\end{figure}

\section{Auto-correlation of Hidden Channels}

The high-dimensional nature of embedding vectors and hidden state channels in typical LLMs makes it challenging to comprehend their functionality. In this study, we employed a non-traditional auto-correlation calculation to reveal some characteristics of these channels. Specifically, we used the following formula: $C_{auto} = \sum_{i=1}^{128\times156\times8} A_{ij} \times A_{(i+d)k}$, where the second amplitude index is shifted by a variable amount d, which we refer to as the state propagation delay. As our sequence is finite, we padded any mismatches with zeros.

It is important to note that the data for the sequence is not exactly one continuous time sequence since we performed multiple snapshots under different context tokens, and each snapshot corresponds to only one token generation. Additionally, as we push the delay, there are more and more correlations between different snapshots, even though the corresponding channel remains the same. Nevertheless, this new formula enabled us to identify the functionalities of some of the channels.

Fig. 4 shows two plots generated by sorting the auto-correlation function in ascending order based on their value at period 512. Surprisingly, we discovered many beautiful patterns within the noise-like wave patterns. Previously, we were unsure of where the position embedding information went in the hidden states, which always begin as a waveform composed of token embedding vectors and position embedding vectors. However, with this visualization of the autocorrelation, we can observe that the position encoding primarily goes into some of the channels, as evidenced by the waves with different harmonics of 128, which is the context size of the network, as shown in Fig.4(a). We also discovered many nice periodic patterns in a significant range in Fig.4(b). Typically, this type of periodic structure indicates that there is a corresponding periodic structure underlying the original oscillation amplitudes. While exploring these wave patterns deserves further research, we defer it for future investigation. Our speculation is that it may be related to the generation of long-range language patterns.

\begin{figure}[htbp]
  \centering
  \includegraphics[width=0.7\linewidth]{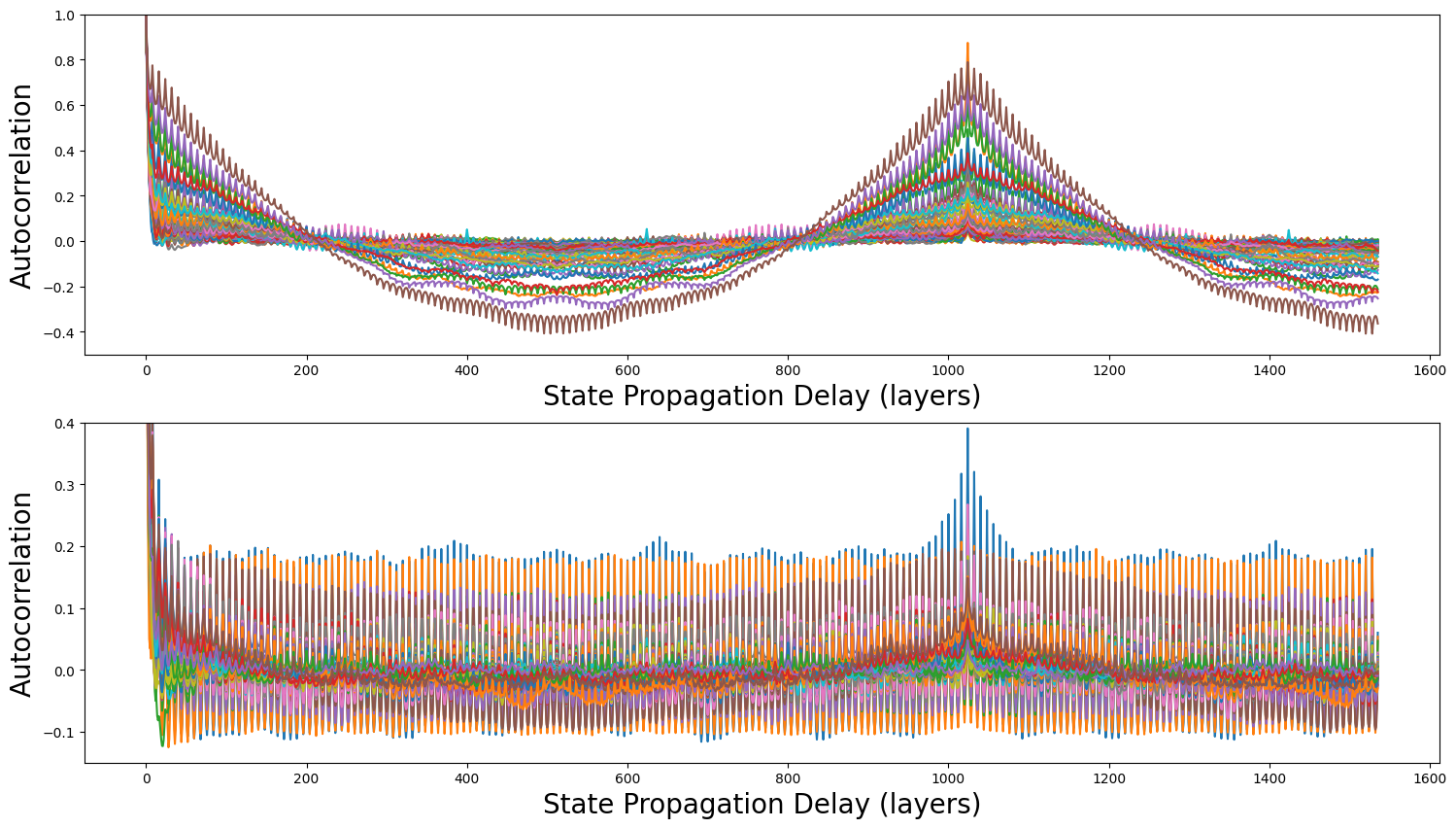}
  \caption{Autocorrelation of Hidden Channels. The autocorrelation time series of the 192 channels are sorted by their values at time delay = 511 layers in ascending order. The first 96 curves are plotted in (a) and the remaining 96 are plotted in (b).}
  \label{fig4}
\end{figure}

Understanding the relationship between the large number of hidden state channels in even a small GPT model containing 192 dimensions can be challenging to comprehend visually. To gain a better understanding of the similarity between the autocorrelation curves of these channels, we transformed each curve into a vector and used t-distributed stochastic neighbor embedding (TSNE) clustering to project the autocorrelation of these channels from their high-dimensional manifolds onto a 2D plane, as shown in Fig. 5(a) and (b). The clustering results revealed remarkable continuity and segmentation among the channels. As we know, similar channel oscillations tend to produce long-range interference patterns, thus controlling long language pattern formations. In addition to the clustering results, we also made additional plots for models trained at different levels in Fig. 5(c)-(f) to illustrate how these manifolds were built up during model training. This kind of manifold visualization can be very useful once we identify the function of each channel. For example, we could use it with reinforcement learning to guide the training of new models with selected channel functionalities, potentially enabling us to create smaller, function-specific models out of LLMs.

\begin{figure}[htbp]
  \centering
  \includegraphics[width=0.7\linewidth]{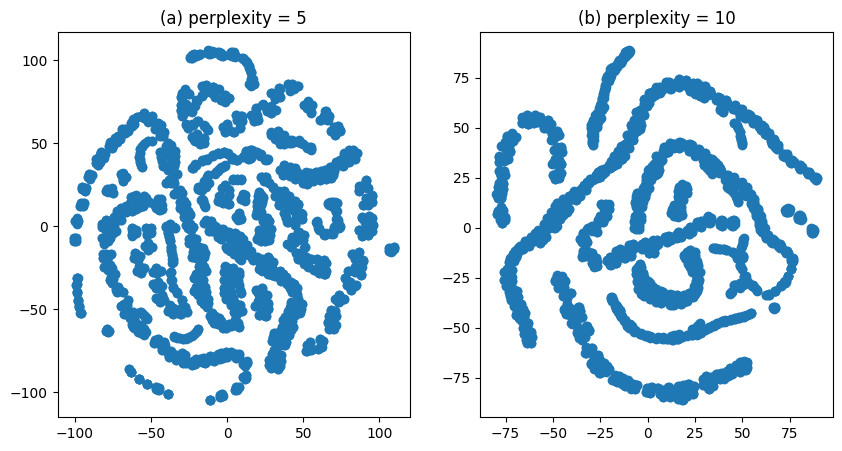}
  \includegraphics[width=0.7\linewidth]{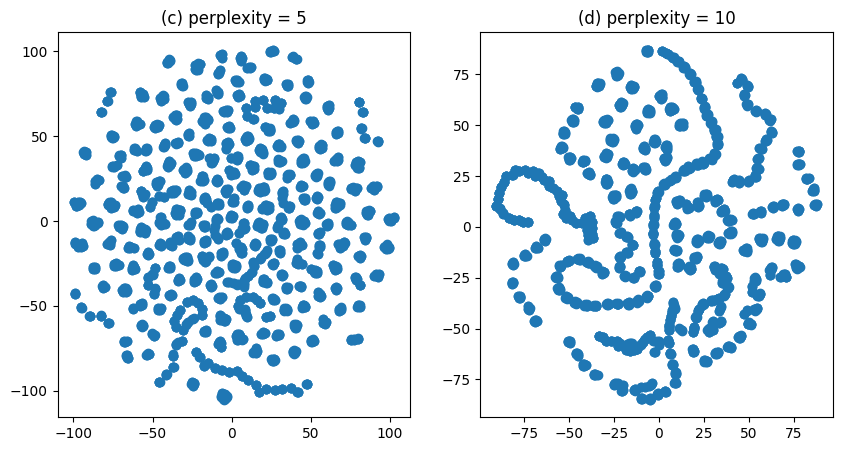}
  \includegraphics[width=0.7\linewidth]{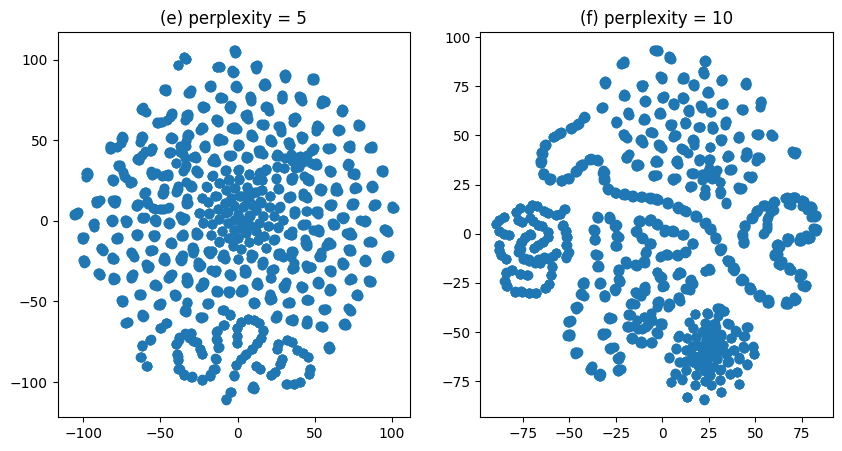}
  \caption{TSNE Projection of Hidden State Channel Autocorrelation Clusters. Each dot corresponds to a hidden state channel, and the clustering and distribution reveals similarities among these channels, as they are projected onto a 2D plane from high-dimensional manifolds. The model was trained at different levels. According to table~\ref{tab:model_training_levels}, condition 1 for (a) and (b), condition 2 for (c) and (d), and condition 3 for (e) and (f). }
  \label{fig5}
\end{figure}

\section{Word Spelling Defects and Phase Transition}

When using a generative model, it is reasonable to expect that the stochastic process involved in token generation may produce incoherent language patterns, such as words with spelling errors that were not present in the training data. This can result in cascading defects or increasing deviations. However, our experiments demonstrate that a well-trained model can suppress these defects through the coherence that has built up in the system, even if the stochastic process occasionally generates words with spelling errors.

To demonstrate this, we employ Poisson statistics, a method commonly used in quantum physics for particle counting and analysis. We first generated 20,000 tokens continuously and then tokenized the text into words based on spaces or punctuation, building a vocabulary from the training data. Next, we tokenized the text sequence generated by the model into words using the same rule and counted each word as either correct or misspelled, depending on whether it existed in the training data vocabulary.

Figure 6 presents histograms of word counts under three different model training conditions. In a well-trained model, as shown in Fig. 6(a), the distribution of misspelled word counts closely follows the Poissonian distribution, indicating that the occurrence of words with defects is independent of each other. In contrast, the distribution of correct word counts shows a significant deviation that is narrower than the corresponding Poissonian distribution in Fig. 6(b). This is typically referred to as a sub-Poissonian distribution, indicating that there is significant interaction or influence among these correctly spelled words. However, we found a revised case when the model was barely trained, as shown in Fig. 6(e) and (f). For models that are in between the two training conditions, we see both coherence and incoherence attempting to influence the next word generation in their own way, resulting in sub-Poissonian distributions in Fig. 6(c) and (d).

\begin{figure}[htbp]
  \centering
  \includegraphics[width=0.7\linewidth]{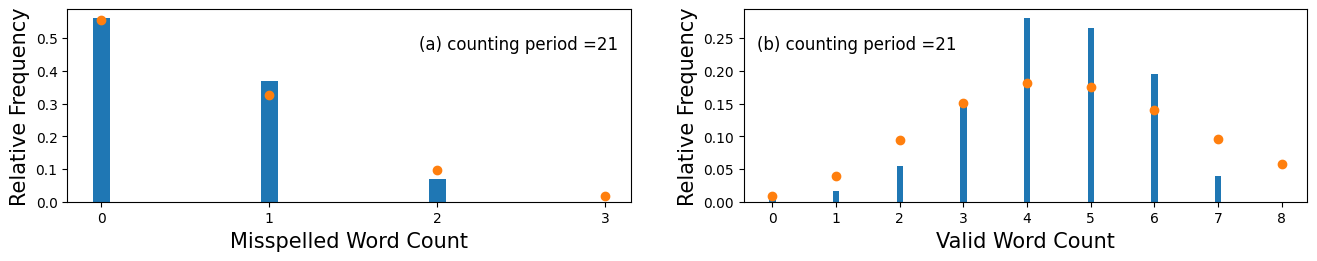}
  \includegraphics[width=0.7\linewidth]{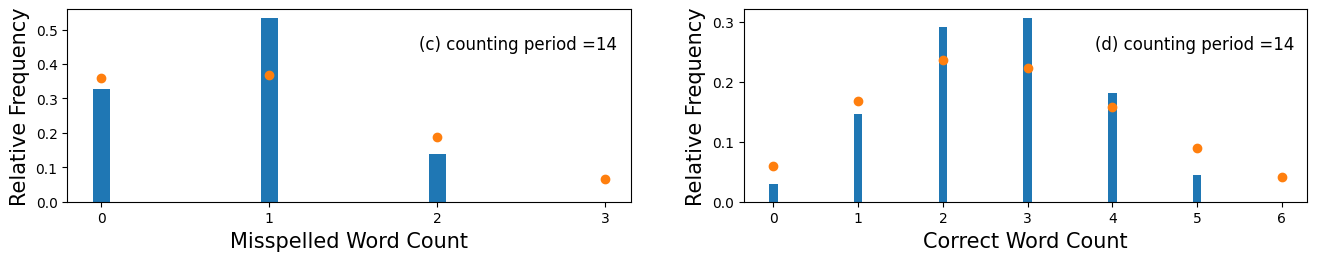}
  \includegraphics[width=0.7\linewidth]{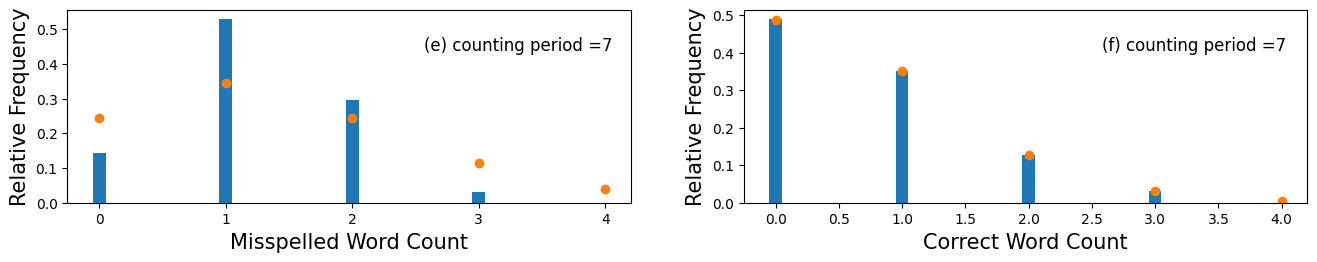}
  \caption{Histograms of Word Counts. (a), (c), and (e) show the distribution of misspelled word counts, while (b), (d), and (f) show the distribution of correct word counts. The bars represent the data, and the dots represent the fit with Poissonian distributions. The model was trained at different levels, as indicated by the conditions in Table~\ref{tab:model_training_levels}: condition 1 for (a) and (b), condition 2 for (c) and (d), and condition 3 for (e) and (f). The y-axis denotes the relative frequency of occurrence, and the x-axis denotes the word count per observation window. The word counting periods range from 21 to 7 character tokens as labeled. These periods are purposely chosen to be equal to or close to the average character token distance between two adjacent misspelled words in the long sequence we observed under the corresponding condition, so that any interactions become more evident.}
  \label{fig6}
\end{figure}

We examine the ratio of misspelled to correct words in a sequence of observation windows of 21 character tokens, as shown in Fig. 7, to gain a more direct view of the word generation process. Even when the model is well-trained, words with defects are generated sporadically, and sometimes at a higher rate within the observation window. However, the rate does not increase in the following window, indicating that the number of defective words does not amplify, as shown in Fig. 7(a). On the other hand, in Fig. 7(b) and (c), when the model is not as well-trained, there are more competitions between the correct words and defects generation, resulting in more fluctuations between the two. This behavior is consistent with the Poisson statistics and indicates that the coherent process becomes strong enough to effectively suppress spelling defects once the model is adequately trained, preventing their cascade amplification. While word generation is a shorter-range language coherence than reasoning, the approach we use can be extended to larger language models and allow for the examination of much longer coherent patterns and the interpretation of their emergent capabilities.

\begin{figure}[htbp]
  \centering
  \includegraphics[width=0.7\linewidth]{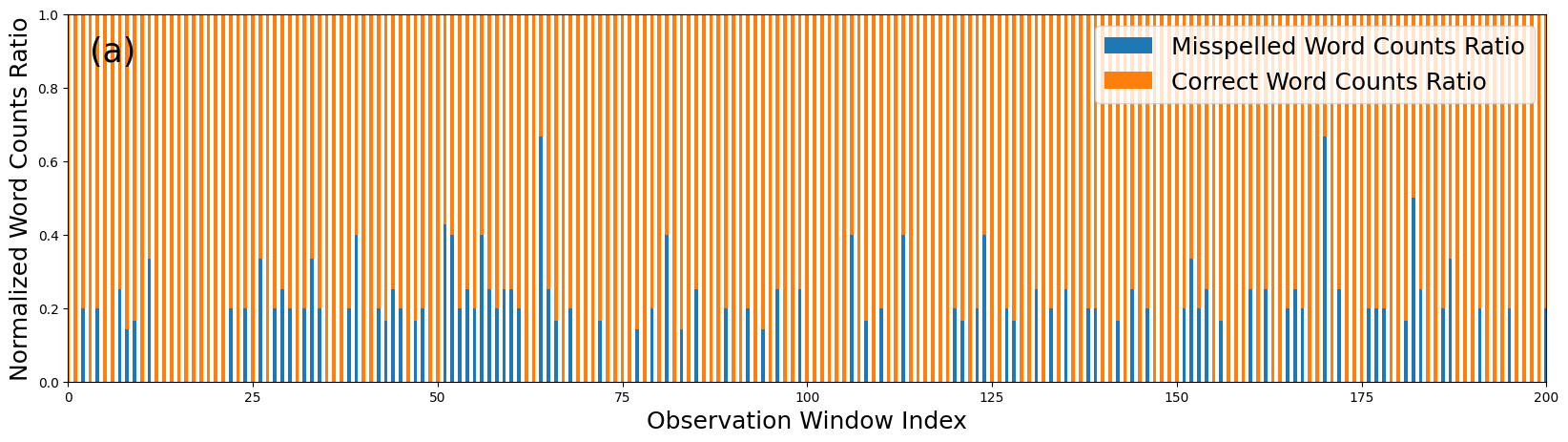}
  \includegraphics[width=0.7\linewidth]{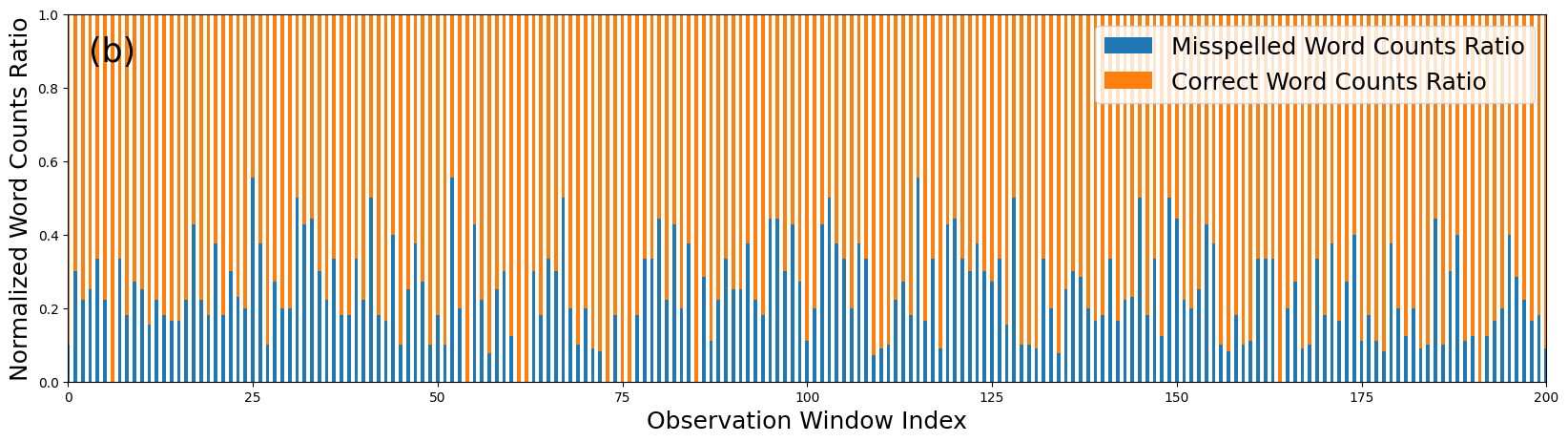}
  \includegraphics[width=0.7\linewidth]{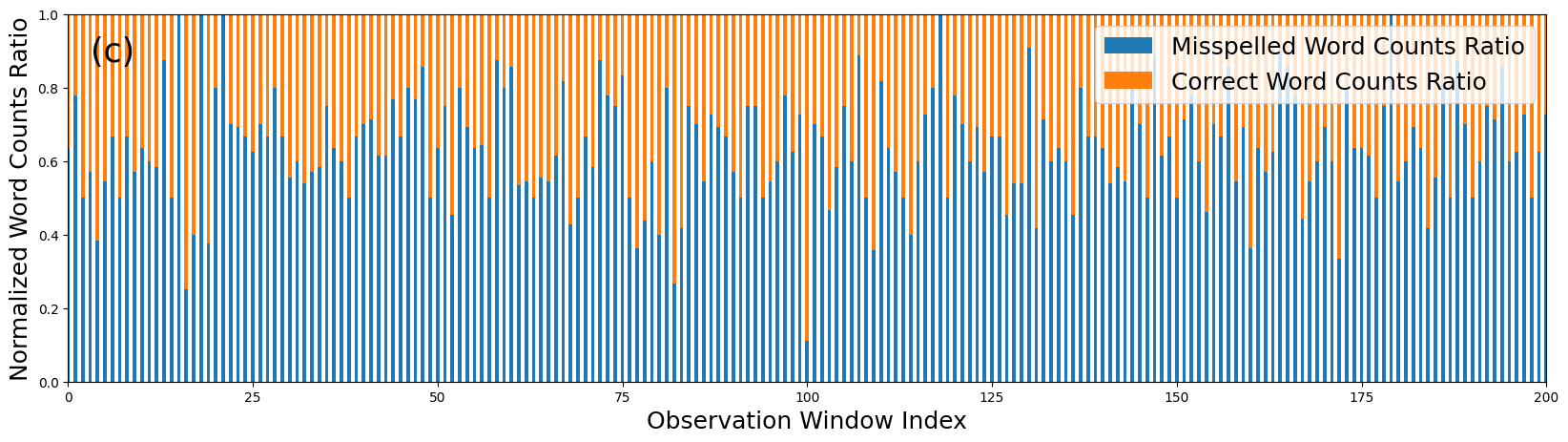}
  
  \caption{Ratio of misspelled to correct words counted in observation windows of 21 tokens. A word is counted as misspelled if its first character appears in the window, but the entire word is not in the training dataset's vocabulary. Conversely, a word is counted as correct if its entire form appears in the vocabulary. The model was trained at different levels, with (a) corresponding to condition 1, (b) corresponding to condition 2, and (c) corresponding to condition 3, as described in Table~\ref{tab:model_training_levels}.}
  \label{fig7}
\end{figure}

\section{Conclusion}

In conclusion, this study delved into the hidden state and channel wave dynamics of a small GPT to gain a better understanding of its inner mechanisms. The coherence of wave patterns was analyzed in terms of cross-channel correlation and individual auto-correlation, which revealed stable and repeatable intrinsic oscillation modes, along with context-aware plasticity and expressiveness in language generation. By analyzing wave patterns, coherence, and clustering, this study offers a systematic way to identify and interpret the functionality of each component, paving the way to understand and control higher-level language pattern formation. Moreover, this study investigated the Poisson statistics of spelling errors in text sequence generation across various levels of model training, revealing a phase-transition-like process as coherence builds up. Once the model is adequately trained, the coherent process becomes strong enough to effectively suppress spelling errors, preventing the cascade amplification of defects. This study has leveraged concepts and techniques from quantum physics to gain novel insights into the dynamics of the small GPT, which can be extended to larger language models that exhibit more complex coherent language patterns, opening up opportunities to interpret their emergent capabilities and develop more specialized models.

\section{Acknowledgement}
We are grateful to Jiandong Xu for fruitful comments and inspiration.


\end{document}